\pdfoutput=1

\documentclass[11pt]{article}

\usepackage{ACL2023}

\usepackage{times}
\usepackage{latexsym}
\usepackage{makecell}
\usepackage{multirow} %
\usepackage[T1]{fontenc}

\usepackage[utf8]{inputenc}

\usepackage{microtype}

\usepackage{inconsolata}

\usepackage{booktabs} 

 \usepackage{graphicx} 
 \usepackage{subfig}
 \usepackage{amsfonts}

\title{Enhancing Fact Retrieval in PLMs through Truthfulness}

\author{Paul Youssef\textsuperscript{$\dag$}  Jörg Schlötterer\textsuperscript{$\dag\ddag$} Christin Seifert\textsuperscript{$\dag$}\\
 \textsuperscript{$\dag$}University of Marburg, \textsuperscript{$\ddag$}University of Mannheim\\
  \texttt{\{paul.youssef, joerg.schloetterer, christin.seifert\}@uni-marburg.de}\\}

\begin{document}
\maketitle
\begin{abstract}
Pre-trained Language Models (PLMs) encode various facts about the world at their pre-training phase as they are trained to predict the next or missing word in a sentence. There has a been an interest in quantifying and improving the amount of facts that can be extracted from PLMs, as they have been envisioned to act as soft knowledge bases, which can be queried in natural language. Different approaches exist to enhance fact retrieval from PLM. Recent work shows that the hidden states of PLMs can be leveraged to determine the truthfulness of the PLMs' inputs. Leveraging this finding to improve factual knowledge retrieval remains unexplored. In this work, we investigate the use of a helper model to improve fact retrieval. The helper model assesses the truthfulness of an input based on the corresponding hidden states representations from the PLMs. We evaluate this approach on several masked PLMs and show that it enhances fact retrieval by up to 33\%. Our findings highlight the potential of hidden states representations from PLMs in improving their factual knowledge retrieval. 

\end{abstract}

\section{Introduction}

Pre-trained Language Models (PLMs) absorb numerous facts about the world from their pre-training data~\cite{petroni-etal-2019-language, roberts-etal-2020-much}. This has sparked the interest of the NLP community in examining and improving the amount of knowledge that can be extracted from PLMs~\cite{youssef-etal-2023-give}. Indeed, several enhancements have been proposed, which go beyond manual prompts~\cite{petroni-etal-2019-language} to improve fact retrieval by directly optimizing prompts~\cite{shin-etal-2020-AUTOPROMPT, zhong-etal-2021-factual, li-etal-2022-spe}, re-writing prompts with other PLMs~\cite{haviv-etal-2021-bertese, zhang-etal-2022-promptgen}, finetuning the PLMs themselves~\cite{roberts-etal-2020-much, fichtel-etal-2021-prompt} or debiasing the outputs of PLMs~\cite{zhao-etal-2021-calibrate, dong-etal-2022-calibrating, wang-etal-2023-towards-alleviating}.

\begin{figure}[ht!]

\includegraphics[width=\columnwidth, trim={2cm 10cm 2cm 1cm}, clip]{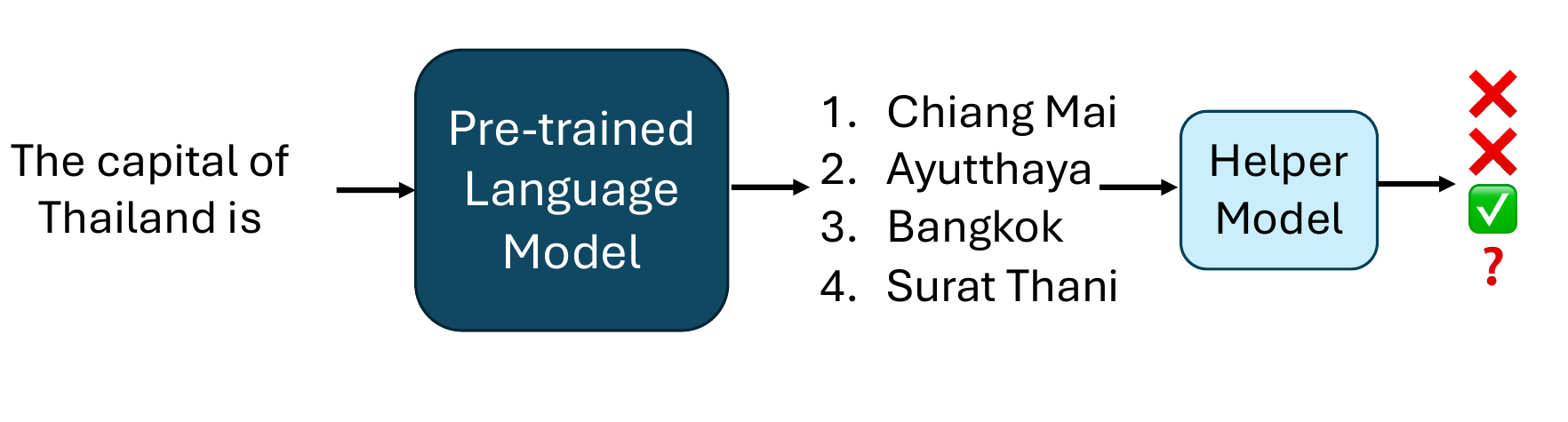}

\caption{An overview of our method for factual knowledge retrieval. A helper model decides which of the proposed answers is correct based on  hidden state representations of the answer from the probed PLM. }
\label{fig:overview}
\end{figure}

Recent work~\cite{burns-etal-2022-discovering} shows that representations from PLMs can be leveraged to determine if the provided inputs are truthful or not, i.e., these representations can be utilized to answer yes/no questions, or to conduct binary classification in an unsupervised manner. However, the utility of these representations for improving fact retrieval has not been examined yet. In this work, we close this gap by investigating, how using a \emph{helper} model that classifies which of the top-$k$ retrieved answers is correct based on the corresponding representations improves fact retrieval. Figure~\ref{fig:overview} demonstrate an overview of our approach. Our results show an improved performance on several masked PLMs.

In summary, our contributions are the following: i) We investigate the use of a helper model in improving fact retrieval based on hidden representations from PLMs; ii) We show that our approach improves fact retrieval performance by up to 33\%; iii) We analyze how increasing the number of the considered predictions affects the final performance.

\section{Related Work}
\paragraph{Fact retrieval.} Despite the incoherence of factual knowledge in PLMs~\cite{youssef-etal-2024-queen}, many works exist that aim to improve fact retrieval from PLMs. One improvement direction is the prompts used to retrieve facts, which have undergone many refinements. After the use of manual prompts by~\cite{petroni-etal-2019-language}, the focus shifted on optimizing prompts either through automatically finding paraphrases that perform better~\cite{qin-eisner-2021-learning}, by optimizing the prompts in discrete space~\cite{shin-etal-2020-AUTOPROMPT} or in a continuous space~\cite{zhong-etal-2021-factual}, or re-writing the prompts by other PLMs~\cite{haviv-etal-2021-bertese, zhang-etal-2022-promptgen}. Another direction for improvement has been the PLMs themselves, which have been finetuned for better fact retrieval~\cite{roberts-etal-2020-much, fichtel-etal-2021-prompt} or to become more robust to changes in the prompts~\cite{elazar-etal-2021-measuring, newman2022padapters}. Other works have focused on debiasing the outputs from PLMs in different ways~\cite{zhao-etal-2021-calibrate, dong-etal-2022-calibrating, malkin-etal-2022-coherence, wang-etal-2023-towards-alleviating, yoshikawa-okazaki-2023-selective}. Our work also aims to improve the outputs from PLMs by leveraging information about the truthfulness of the inputs, which can be derived from the hidden states. For a comprehensive review about factual knowledge retrieval from PLMs, we refer the interested reader to~\cite{youssef-etal-2023-give}.

\paragraph{Truthfulness in Language Models.} \citet{burns-etal-2022-discovering} show that the hidden states from LLMs can be used in an unsupervised learning setting to distinguish between truthful and untruthful statements. Similarly, \citet{azaria-mitchell-2023-internal} leverage the hidden states of LLMs to train a feedforward neural network to predict the truthfulness of the LLMs' inputs, and show its effectiveness on several LLMs.  \citet{pacchiardi2024how} show that it is possible to detect untruthful answers from LLMs with no access to the hidden states from the LLMs by asking several simple yes/no questions and feeding the LLMs' outputs into a logistic regression model. Despite its simplicity, their approach is shown to generalize to different architectures and LLMs that are finetuned to output lies. In summary, many works exist that show that the hidden states of LLMs can be leveraged to predict the truthfulness of their inputs. In this work, we leverage the truthfulness signal in hidden states to improve fact retrieval.

\section{Methodology}

Factual knowledge in PLMs is estimated by evaluating how often can PLMs correctly
predict an object entity $o$, given a subject entity $s$ and a relation $r$ that are expressed through a prompt $p(s,r)$. Given a PLM $\mathcal{M}$, its predicted object $\hat{o} = argmax_o \ \mathbb{P}_{\mathcal{M}, p(s,r)} [o]$ is the token with the highest probability given the prompt that contains the subject and the relation $p(s,r)$.

In this work, we consider top-$k$ outputs from $\mathcal{M}$ instead of using the top-1 prediction $\hat{o}$. In order, to decide which of the $k$ outputs is the final prediction, we leverage a helper model $\mathcal{H}$. $\mathcal{H}$ takes as input the hidden state that corresponds to the final token in the input from the last encoder layer of the PLM $\mathcal{M}$ after inputting the prompt $p(s,r,o_i)$, where $o_i$ refers to the top $i$-th prediction and $i \in \{1,k\}$. Assuming that $p(s,r,o_i)$ consists of $j$ tokens and $\mathcal{M}$ has $l$ layers, we use the hidden state $h_{j,l}$ from $\mathcal{M}(p(s,r,o_i))$. $h_{j,l}$ represents the hidden state that corresponds to the $j$-token from the $l$-th layer, as input to the helper model $\mathcal{H}$. $\mathcal{H}$ classifies $h_{j,l}$ as either truthful or not. Since \citet{burns-etal-2022-discovering} show that the hidden states of PLMs contain information about the truthfulness of their inputs  we expect the helper model $\mathcal{H}$ to positively affect the factual knowledge retrieval performance.

\section{Experiments}
Here, we describe the data and PLMs, for our experiments in detail.

\subsection{Datasets and PLMs}

\paragraph{Datasets.} we use 2 test sets in our experiments in order to evaluate the fact retrieval performance: 
\begin{itemize}
    \item \textbf{LAMA}: we use the T-REx~\cite{elsahar-etal-2018-rex} subset of LAMA~\cite{petroni-etal-2019-language}, which is often used to estimate factual knowledge in PLMs.
    \item \textbf{WIKIUNI}: A second dataset for estimating factual knowledge in PLMs \cite{cao-etal-2021-knowledgeable}. In contrast to LAMA,  in this dataset the ground truth objects are uniformly distributed. 
\end{itemize}

To train the helper model $\mathcal{H}$, we consider two training sets: 
\begin{itemize}
    \item \textbf{AUTOPROMPT}: The dataset used by \citet{shin-etal-2020-AUTOPROMPT} to optimize prompts in a discrete space.
    \item \textbf{WIKIUNI}: The same as the dataset mentioned above for testing. 
\end{itemize}

\paragraph{Training the helper model.} since the training sets contain only the ground truth object, we sample \emph{untruthful} examples from $\mathcal{M}$'s outputs to train $\mathcal{M}$. The untruthful examples correspond to the top $k+1$ outputs. We report the optimal top $k+1$ for each setting, and refer to this as the \textbf{Neg. Index}. We also report the accuracy of the helper model. As helper model, we use a simple logistic regression model with L1 regularization. Since the frequency distribution of the object entities between the training and test sets might have an impact on the performance, we report the Pearson correlation coefficient \textbf{Corr} between these two distribution.%

\paragraph{PLMs.} we experiment with four models: BERT-base, BERT-large~\cite{devlin-etal-2019-bert}, T5-base and T5-large~\cite{raffel-etal-2020-t5}. We summarize the number of parameters and architecture for each model in Table~\ref{tab:models}. For BERT, we exclude examples with objects that consist of more than one token. For T5 model, we keep both one-token and multiple-token objects, and use Typed Querying (TyQ)~\cite{kassner-etal-2021-multilingual} to extract the top-$k$ predictions. In TyQ, the number objects to be considered is limited to a subset in contrast to the normal case where the whole vocabulary is considered. TyQ makes it easier to consider objects consisting of multiple tokens. We augment the set of objects for each relation with predictions from BERT. We replace the subject with \texttt{NA} to extract the predictions, keeping only the relation in the prompt. As a baseline, we use the object with the highest probability from the PLMs. Following ~\citet{burns-etal-2022-discovering}, we extract the hidden representations from the last encoder layer for both model types.

\begin{table}
\centering
\begin{tabular}{lll}
\hline
\textbf{Model} & \textbf{\#Parameters} & \textbf{Architecture}\\
\hline
BERT-base & 110M & encoder-only \\
BERT-large & 345M & encoder-only \\
T5-base & 220M & encoder-decoder \\
T5-large & 770M & encoder-decoder \\

\hline
\end{tabular}
\caption{Models with number of parameters and architectures.}
\label{tab:models}
\end{table}

\subsection{Results \& Discussions}

\begin{table*}[!htbp]
    \centering

\begin{tabular}{l l c c c c c c c}

\toprule
\makecell{\textbf{Training} \\ \textbf{Set}} & \makecell{\textbf{Test} \\ \textbf{Set}} & \textbf{Model} & \textbf{Baseline} & \textbf{Ours} & \textbf{Diff.\%} & \textbf{Corr} & \makecell{\textbf{Neg.} \\ \textbf{Index}} & \makecell{\textbf{Helper}\\ \textbf{Acc.}} \\
\midrule

\multirow{8}{*}{\rotatebox{90}{AUTOPROMPT}} & \multirow{4}{*}{\rotatebox{90}{LAMA}} 
 & BERT-base & 29.29 & 38.99 & \textbf{33.12} & 0.74 & 21 & 87.33 \\
 & & BERT-large & \textbf{31.37} & 40.04 & 27.64 & 0.74 & 21 & 87.54 \\
 & & T5-base & 14.52 & 16.59 & 14.26 & 0.74 & 31 & 89.22 \\
 & & T5-large & 17.63 & 19.50 & 10.61 & 0.74 & 101 & 88.09 \\
\cmidrule{2-9}
 & \multirow{4}{*}{\rotatebox{90}{WIKIUNI}}
 & BERT-base & 14.94 & 16.93 & 13.32 & 0.04 & 91 & 90.30 \\
 & & BERT-large & \textbf{16.79} & 19.14 & 14.00 & 0.04 & 81 & 90.91 \\
 & & T5-base & 5.90 & 6.15 & 4.24 & 0.04 & 21 & 88.20 \\
 & & T5-large & 7.44 & 7.43 & -0.13 & 0.04 & 11 & 89.45 \\
\midrule

\multirow{4}{*}{\rotatebox{90}{WIKIUNI}} & \multirow{4}{*}{\rotatebox{90}{LAMA}}
 & BERT-base & 29.29 & 34.23 & \textbf{16.87} & -0.03 & 101 & 82.53 \\
 & & BERT-large & \textbf{31.37} & 36.26 & 15.59 & -0.03 & 71 & 81.39 \\
 & & T5-base & 14.53 & 16.21 & 11.56 & -0.03 & 101 & 79.82 \\
 & & T5-large & 17.63 & 18.94 & 7.43 & -0.03 & 101 & 79.08 \\
\bottomrule
\end{tabular}
    \caption{Factual knowledge retrieval performance (accuracy) on several test sets. Training set is only used with helper approach. \textbf{Diff.\%} refers to the percentage improvement in performance. }
    \label{tab:results}
\end{table*}

Table~\ref{tab:results} shows the results of our experiments. In general, the top-1 accuracy improves when the helper model is used. The improvements vary between more than 33\% (BERT with AUTOPROMPT and LAMA) and 4\% (T5-base with AUTOPROMPT and WIKIUNI). One exception is T5-large with AUTOPROMPT and WIKIUNI, where the performance does not change significantly. 

Even though the improvements in performance are the highest when the correlation between the training and test sets is high (AUTOPROMPT and LAMA), we still notice improvements when there is no correlation between the training and test sets (WIKIUNI and LAMA). This verifies the findings of \citet{burns-etal-2022-discovering} that the hidden states contain information about the truthfulness of the inputs. This is further verified by the high accuracy of the helper model (> 80\% in most cases). We also notice that the fact retrieval performance and the gains in performance after using a helper model are higher for BERT-models than for T5-models. This shows that encoder-only models are not only better at fact retrieval~\cite{lewis-etal-2020-bart, zhang-etal-2021-need, youssef-etal-2023-give}, but also their hidden states contain more information about the truthfulness of the inputs.

\begin{figure}%
\centering
\subfloat[][BERT-models]{%
  \includegraphics[width=0.99\columnwidth, trim={1cm 0cm 0cm 0cm}, clip]{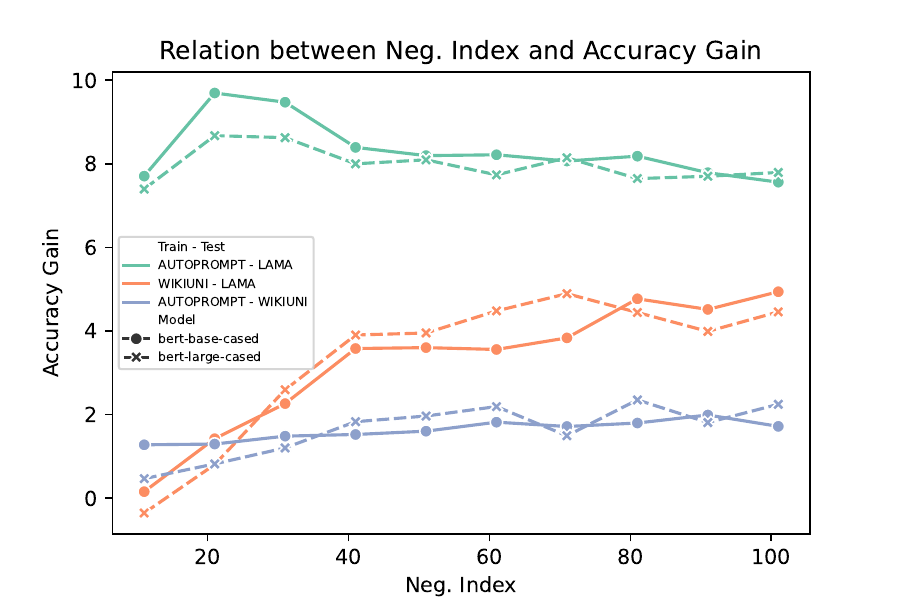}%
  \label{fig:negindex:a}
} 

\subfloat[][T5-models]{%
  \includegraphics[width=0.99\columnwidth, trim={1cm 0cm 0cm 0cm}, clip]{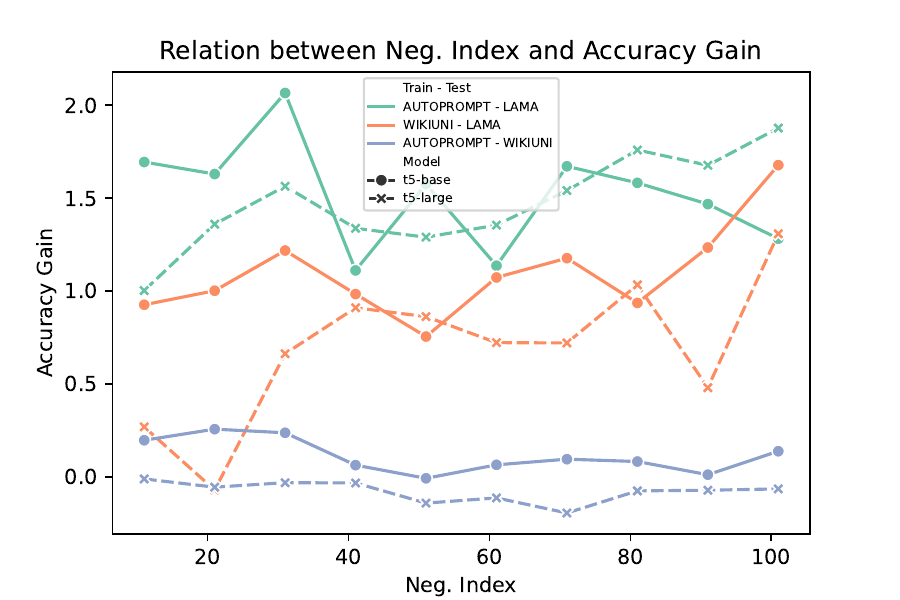}%
  \label{fig:negindex:b}
} 

\caption{Relation between \textbf{Neg. Index (K+1)} and Gain in Accuracy. Top: BERT-models, bottom: T5-models}
\label{fig:negindex}
\end{figure}

\paragraph{Effect of Neg. Index ($k+1$)}We also investigate the relation between the \textbf{Neg. Index} and the gain in performance. Note that as we increase the Neg. Index, we also increase the number of predictions $k$ to be considered by the helper model. For example, a Neg. Index of 21 corresponds to a $k=20$, i.e., the helper model assesses the top $k=20$ predictions, and returns the first one that is predicted to be truthful. 

Figure~\ref{fig:negindex}  shows the relation between Neg. Index ($k+1$) and gains in performance for BERT models~\ref{fig:negindex:a}  and T5 models~\ref{fig:negindex:b}. We notice that the gains in performance for BERT-base and BERT-large look similar. The gains on AUTOPROMPT-LAMA reach their highest at 21, and start dropping slightly after that. On WIKIUNI-LAMA, the gains steadily increase with Neg. Index and rich their highest at 71 (BERT-large) and 101 (BERT-base). A similar trend can be noticed on AUTOPROMPT-WIKIUNI. In general, one can notice that \emph{when there is a high correlation between the train and test sets (e.g., AUTOPROMPT-LAMA) then the gains in performance are high and are attained at small Neg. Index ($k$) values. Conversely, when there is no correlation between the train and test sets the gains in performance are lower and are reached at larger Neg. Index ($k$) values.} This can either be attributed to the availability of more predictions to choose from (with the increase of $k$), or to a potential improvement in the accuracy of the helper model. A further investigation is needed to disentangle the effect of both factors. 

The gains in performance for T5-models are smaller (see Figure~\ref{fig:negindex} b) than those of BERT-models. Here, we notice that the gains on WIKIUNI are the smallest and do not vary much. For T5-large, the performance even degrades slightly. On LAMA, there is more variance, and we notice that when there is no correlation between the training and test sets (WIKIUNI-LAMA) the best performance is reached at high Neg. Index values ($101$ for both T5-base and T5-large). When the correlation is high (AUTOPROMPT-LAMA) the best performance is reached at lower values (e.g., T5-base at Neg. Index$ = 31$ ). An exception here is T5-large where the best performance is reached at Neg. Index $ = 101$. However, the difference is not big compared to a peak at a lower Neg. Index $= 31$ (1.56 vs. 1.87). We believe the differences in performance between BERT and T5 can be attributed to their different architectures.

\section{Conclusion}
In this work, we investigated the use of a helper model to improve fact retrieval. The helper model relies on the hidden state representations from PLMs to determine the truthfulness of the corresponding inputs. We showed the effectiveness of this approach in improving fact retrieval on several masked PLMs. Furthermore, we showed that increasing the number of the considered predictions affects the performance positively, especially in cases where the answer frequencies between the training and test sets are not correlated.
Even though our approach for improving fact retrieval leads to an improved performance. It is nonetheless more computationally demanding, since it requires extracting the hidden states for all potential predictions. In future work, we aim to optimize this method and extend our evaluation to LLMs.

\section{Limitations}
Our experiments do not include any LLMs due to the high computational costs associated with LLMs under our approach.

\bibliography{anthology,custom}
\bibliographystyle{acl_natbib}

\end{document}